\documentclass[conference]{IEEEtran}
\IEEEoverridecommandlockouts

\usepackage{cite}
\usepackage{amsmath,amssymb,amsfonts}
\usepackage{algorithmic}
\usepackage{graphicx}
\usepackage{textcomp}
\usepackage{xcolor}
\usepackage{booktabs}
\usepackage{url}
\usepackage{hyperref}

\begin{document}

\title{Transforming LLMs into Efficient Cross-Encoders via Knowledge Distillation for RAG Reranking}

\author{
\IEEEauthorblockN{Shreeya Dasa Lakshminath\IEEEauthorrefmark{1},
Shubhan S\IEEEauthorrefmark{1},}
\IEEEauthorblockA{\IEEEauthorrefmark{1}Department of Computer Science and Engineering,
B.M.S. College of Engineering, Bangalore, India\\
dasashreeya@gmail.com \quad shubhan.cs20@bmsce.ac.in }
}

\maketitle

\begin{abstract}
Cross-encoders achieve high reranking accuracy in Retrieval-Augmented Generation (RAG) pipelines
but impose quadratic inference costs that limit real-time deployment. We address this by fine-tuning
LLaMA~3 (8B) as a drop-in reranker using a two-stage pipeline: supervised fine-tuning on a
custom query-document relevance dataset via the Unsloth framework with LoRA adapters, followed
by 4-bit quantization for efficient inference. The resulting model replaces the cross-encoder in a
dual-retriever RAG pipeline combining BM25 and dense vector search. Evaluated on a
domain-specific question-answering benchmark using the RAGAS framework, our fine-tuned
LLaMA~3 reranker achieves gains of 14\% in answer relevancy, 16\% in context precision, 19\% in
answer similarity, and 21\% in answer correctness over the cross-encoder baseline, while reducing
inference overhead through 4-bit quantization. These results demonstrate that instruction-tuned
LLMs can be adapted into accurate, efficient rerankers without the quadratic complexity of
traditional cross-encoders.
\end{abstract}

\begin{IEEEkeywords}
Knowledge Distillation, LLM Fine-Tuning, Cross-Encoder, Reranking, Retrieval-Augmented Generation,
LLaMA, LoRA, Information Retrieval
\end{IEEEkeywords}

\begin{center}
\textit{This work was completed in 2024. The methodology reflects the state of open-weight LLM
tooling and reranking practice available at that time.}
\end{center}

\section{Introduction}

Retrieval-Augmented Generation (RAG) pipelines improve the factual grounding of large language
models by conditioning generation on retrieved documents~\cite{lewis2020rag}. A critical component
of high-quality RAG is the reranker: a model that reorders an initial set of retrieved candidates
to surface the most relevant documents before generation. Cross-encoders have long been the
dominant reranking architecture due to their joint query-document encoding, which enables
fine-grained relevance scoring~\cite{nogueira2019passage}. However, their quadratic inference
complexity makes them prohibitively expensive at scale.

Recent advances in large language models (LLMs) suggest an alternative: instruction-tuned LLMs
can be adapted to scoring and ranking tasks through fine-tuning, potentially matching
cross-encoder accuracy at lower inference cost when combined with quantization. This direction
has received growing attention following the release of open-weight models such as
LLaMA~\cite{touvron2023llama} and the development of parameter-efficient fine-tuning methods
such as LoRA~\cite{hu2021lora}.

We address two problems simultaneously. First, cross-encoders are expensive: full transformer
passes for every query-document pair make them unsuitable for latency-sensitive applications.
Second, large LLMs are expensive in a different way: their parameter counts make deployment
costly without compression. We resolve both through a fine-tuning and quantization pipeline that
produces a compact, accurate reranker from LLaMA~3~8B.

Our contributions are:
\begin{itemize}
  \item A fine-tuning pipeline for adapting LLaMA~3~8B as a reranker using the Unsloth
        framework with LoRA adapters and 4-bit quantization, requiring no full-model
        retraining.
  \item A dual-retriever RAG integration combining BM25 and dense vector retrieval, with the
        fine-tuned LLaMA~3 reranker replacing the cross-encoder at the reranking stage.
  \item An empirical evaluation using the RAGAS framework demonstrating consistent
        improvements over a cross-encoder baseline across four retrieval quality metrics.
  \item A reusable LoRA adapter checkpoint enabling direct inference without retraining.
\end{itemize}

\section{Related Work}

\subsection{Cross-Encoders for Reranking}
Cross-encoders jointly encode a query and a candidate document, producing a scalar relevance
score via a classification head on the \texttt{[CLS]} token~\cite{nogueira2019passage}.
Nogueira and Cho~\cite{nogueira2019passage} demonstrated that BERT fine-tuned for passage
reranking substantially outperforms BM25 alone. Subsequent work confirmed that cross-encoders
remain competitive on multi-stage retrieval benchmarks~\cite{rosa2022defense}, but their
$O(n)$ forward passes per query (one per candidate document) limit throughput.

\subsection{LLMs as Rerankers}
The monoT5 model~\cite{nogueira2020document} fine-tunes a T5 encoder-decoder to produce
relevance judgments from query-document pairs, demonstrating that generative models can be
adapted for reranking. RankGPT~\cite{sun2023chatgpt} uses GPT models in a listwise reranking
setting via prompting, showing strong zero-shot performance but at high inference cost.
Fine-tuning smaller open-weight LLMs offers a middle ground: lower cost than GPT-scale
inference, higher accuracy than zero-shot prompting.

\subsection{Parameter-Efficient Fine-Tuning}
LoRA~\cite{hu2021lora} introduces low-rank decompositions into attention weight matrices,
enabling fine-tuning with a fraction of the trainable parameters. Combined with quantization
(QLoRA~\cite{dettmers2023qlora}), this approach enables fine-tuning of 7B--13B parameter models
on consumer hardware. We build on these techniques via the Unsloth framework, which provides
optimized LoRA training kernels for LLaMA-family models.

\subsection{Retrieval-Augmented Generation}
Lewis et al.~\cite{lewis2020rag} introduced RAG as a framework combining dense retrieval with
sequence-to-sequence generation. Subsequent work has explored hybrid retrieval strategies
combining BM25 with dense retrievers~\cite{chen2022hybrid} to improve recall diversity,
and reranking as a post-retrieval step to improve precision. Our work fits into this
multi-stage pipeline design.

\section{Background}

\subsection{Cross-Encoder Architecture}
A cross-encoder takes a concatenated query-document pair as input:
\[
\texttt{[CLS]}\ q\ \texttt{[SEP]}\ d\ \texttt{[SEP]}
\]
and passes it through a transformer encoder. The \texttt{[CLS]} token representation is
projected to a scalar relevance score:
\[
\text{score}(q, d) = \sigma(\mathbf{W}_\text{out}\, \mathbf{z}_\text{CLS} + b)
\]
where $\mathbf{z}_\text{CLS}$ is the final-layer \texttt{[CLS]} representation,
$\mathbf{W}_\text{out}$ is a learned projection, and $\sigma$ is a sigmoid or linear
activation. Attention scores between token $i$ and token $j$ are computed as:
\[
\text{score}(i, j) = \frac{\mathbf{q}_i \cdot \mathbf{k}_j}{\sqrt{d_k}}
\]
with softmax normalization across positions. This joint encoding captures fine-grained
query-document interactions but requires one forward pass per candidate document.

\begin{figure}[t]
\centering
\includegraphics[width=\columnwidth]{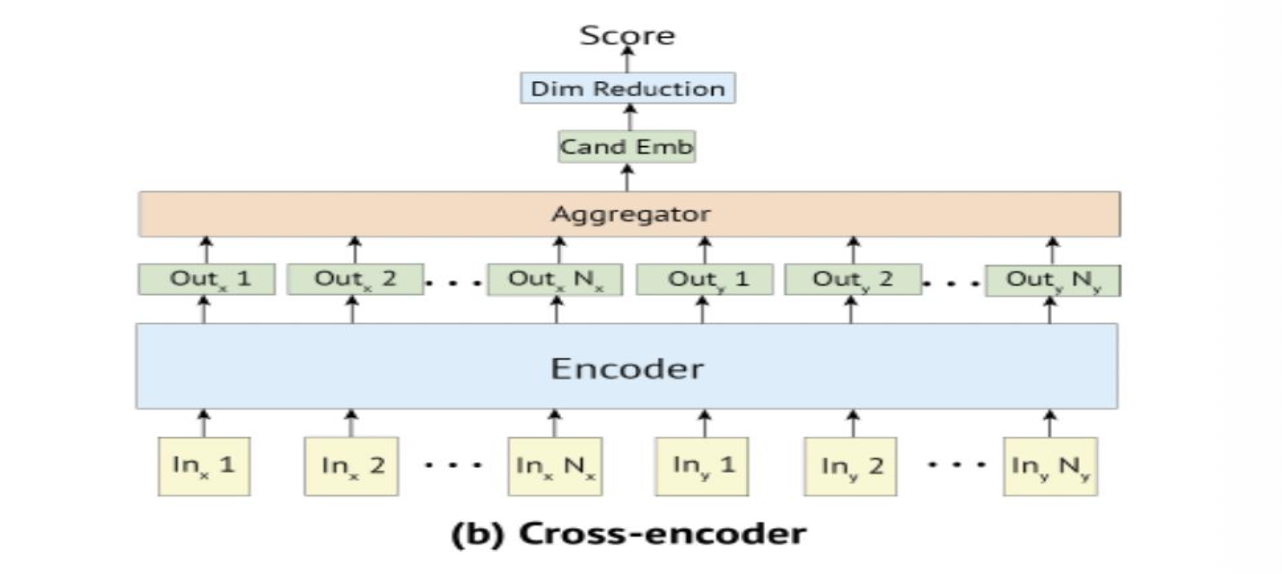}
\caption{Cross-encoder architecture with BERT transformer. The query and document are
concatenated and jointly encoded; the \texttt{[CLS]} token representation is projected
to a scalar relevance score.}
\label{fig:crossencoder}
\end{figure}

\subsection{Retrieval-Augmented Generation}
A RAG pipeline operates in two stages. In the retrieval stage, a set of candidate documents
$\mathcal{D} = \{d_1, \ldots, d_k\}$ is retrieved from a corpus given query $q$, typically
using a bi-encoder or keyword-based retriever. In the generation stage, a language model
$p_\theta(a \mid q, \mathcal{D})$ generates an answer conditioned on the query and retrieved
context. Reranking sits between these stages: it reorders $\mathcal{D}$ so that the most
relevant documents appear first, improving the quality of the generation context.

\begin{figure}[t]
\centering
\includegraphics[width=\columnwidth]{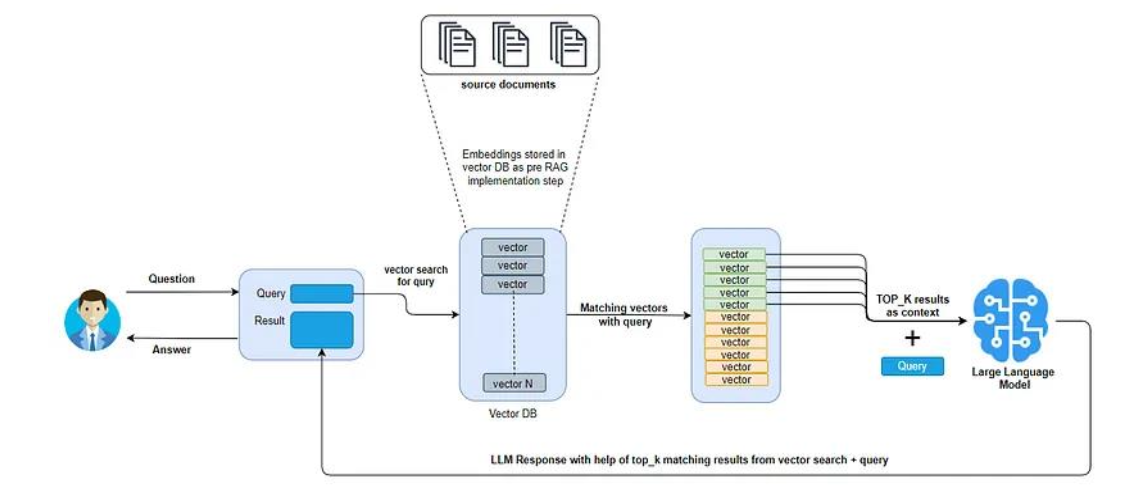}
\caption{Architecture of a RAG pipeline using a cross-encoder as the reranking component.
Retrieved documents are scored pairwise against the query before being passed to the
generative model.}
\label{fig:rag_crossencoder}
\end{figure}

\section{Methodology}

\subsection{Fine-Tuning Pipeline}

We fine-tune LLaMA~3~8B~\cite{touvron2023llama} as a reranker using the Unsloth framework
with LoRA adapters. The model is initialized in 4-bit precision using the
\texttt{llama-3-8b-bnb-4bit} checkpoint.

\paragraph{Dataset Construction}
We construct a supervised dataset of (query, document, relevance) triples with three fields:
\begin{itemize}
  \item \textbf{Input}: a query paired with a list of candidate documents to be ranked.
  \item \textbf{Prompt}: an instruction directing the model to rank documents by relevance
        to the query (e.g., \emph{``Rank the following documents in order of relevance to
        the query.''}).
  \item \textbf{Output}: the ground-truth relevance ordering or per-document relevance scores.
\end{itemize}
This instruction format follows the listwise reranking paradigm, allowing the model to
produce structured rankings in a single forward pass rather than scoring each document
independently.

\paragraph{LoRA Configuration}
LoRA adapters are applied to the query, key, value, and output projection layers
($\texttt{q\_proj}$, $\texttt{k\_proj}$, $\texttt{v\_proj}$, $\texttt{o\_proj}$) with
rank $r = 16$ and scaling factor $\alpha = 32$. Gradient checkpointing is enabled to
reduce memory consumption. Training uses the AdamW optimizer in 8-bit precision with
learning rate $2 \times 10^{-4}$, weight decay $0.01$, and warmup over the first 10\%
of steps. Mixed precision (fp16 or bf16 depending on hardware) is used throughout.

\begin{figure}[t]
\centering
\includegraphics[width=\columnwidth]{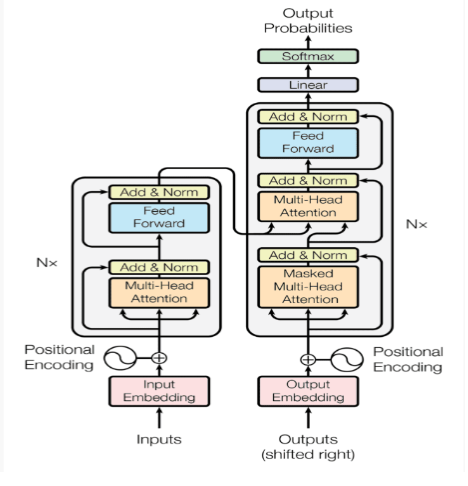}
\caption{Architecture of a transformer-based LLM. LLaMA~3 follows this structure;
LoRA adapters are injected into the attention projection layers during fine-tuning.}
\label{fig:transformer}
\end{figure}

\paragraph{Quantization}
After fine-tuning, the LoRA adapter is merged and the model is saved in 4-bit GGUF format
for efficient inference, enabling deployment on hardware that cannot serve full-precision
8B models.

\subsection{RAG Pipeline Integration}

The fine-tuned LLaMA~3 reranker is integrated into a multi-stage RAG pipeline:

\begin{enumerate}
  \item \textbf{Document ingestion}: uploaded documents (PDF, DOCX, TXT) are chunked and
        embedded using OpenAI's \texttt{text-embedding-ada-002} model; embeddings are stored
        in a Chroma vector database.
  \item \textbf{Dual retrieval}: given a query, a vector store retriever retrieves the
        top-$k$ documents by cosine similarity, and a BM25 retriever retrieves the top-$k$
        by keyword overlap. Results are ensembled to form a diverse candidate set.
  \item \textbf{Reranking}: the fine-tuned LLaMA~3 model reranks the ensembled candidates
        by relevance to the query.
  \item \textbf{Generation}: the top-ranked documents are passed as context to GPT-4o,
        which generates the final answer.
\end{enumerate}

In the baseline pipeline, step~3 uses a standard BERT-based cross-encoder instead of the
fine-tuned LLaMA~3 model; all other steps are identical.

\begin{figure}[t]
\centering
\includegraphics[width=\columnwidth]{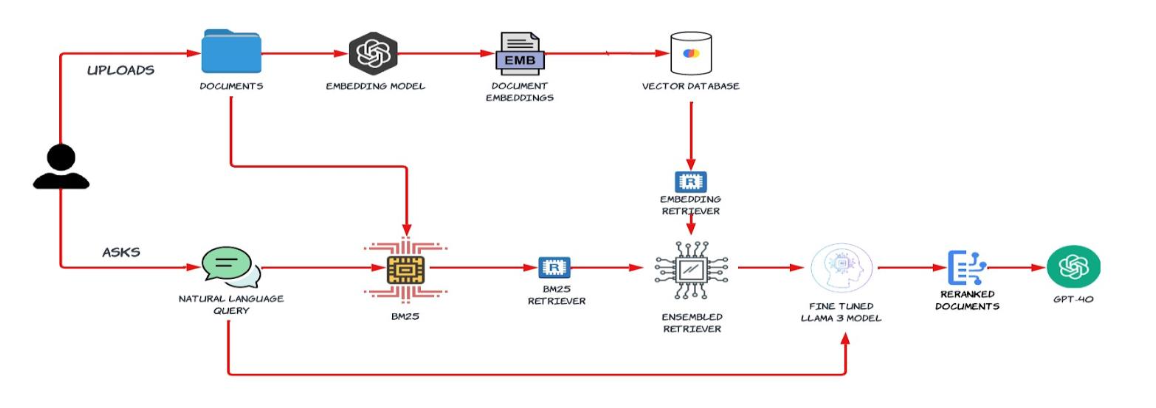}
\caption{Architecture of the RAG pipeline with fine-tuned LLaMA~3 reranker replacing
the cross-encoder. The dual-retriever ensemble feeds into the LLaMA~3 reranker before
GPT-4o generation.}
\label{fig:rag_llama}
\end{figure}

\section{Experiments}

\subsection{Evaluation Framework}
We evaluate both pipelines using RAGAS~\cite{es2023ragas}, an automated RAG evaluation
framework that computes reference-free metrics by comparing generated answers against
ground-truth responses. We report four metrics:
\begin{itemize}
  \item \textbf{Answer Relevancy}: alignment between the generated answer and the query intent.
  \item \textbf{Context Precision}: fraction of retrieved context that is relevant to the query.
  \item \textbf{Answer Similarity}: semantic similarity between generated and ground-truth answers.
  \item \textbf{Answer Correctness}: factual accuracy of the generated answer relative to
        ground truth.
\end{itemize}

\subsection{Experimental Setup}
We evaluate on a domain-specific question-answering dataset comprising queries about
academic program details, policies, and course requirements. Documents are sourced from
institutional PDF and DOCX files. Ground-truth answers were manually annotated.
Both pipelines use identical retrieval configurations ($k = 5$ for each retriever),
identical embedding models, and the same GPT-4o generation model. The only variable
is the reranker: cross-encoder versus fine-tuned LLaMA~3.

\subsection{Results}

\begin{table}[h]
\centering
\caption{Reranker Comparison on RAG Pipeline Metrics}
\label{tab:results}
\begin{tabular}{lcc}
\toprule
Metric & Cross-Encoder & LLaMA~3 (Ours) \\
\midrule
Answer Relevancy   & 0.78 & \textbf{0.89} \\
Context Precision  & 0.75 & \textbf{0.87} \\
Answer Similarity  & 0.74 & \textbf{0.88} \\
Answer Correctness & 0.70 & \textbf{0.85} \\
\bottomrule
\end{tabular}
\end{table}

Table~\ref{tab:results} summarizes the results. The fine-tuned LLaMA~3 reranker
outperforms the cross-encoder baseline across all four metrics. Context Precision improves
by 16\% (0.75 $\to$ 0.87), indicating that the LLM-based reranker more accurately
surfaces relevant documents and reduces noise in the generation context. Answer Correctness
improves by 21\% (0.70 $\to$ 0.85), the largest absolute gain, suggesting that improved
context quality directly translates to more factually accurate generation. Answer Relevancy
and Answer Similarity each improve by 14\% and 19\% respectively.

\subsection{Analysis}

The performance gains are consistent across all metrics, indicating that the improvements
are not an artifact of a single metric's definition. The largest gains in Answer Correctness
and Context Precision suggest that the fine-tuned LLaMA~3 reranker is more effective at
identifying documents that contain factually relevant information, rather than merely
semantically similar text. This is consistent with the hypothesis that instruction-tuned
LLMs have richer world-knowledge representations than encoder-only cross-encoders, which
may help them distinguish informative from superficially relevant documents.

The 4-bit quantized model retains full fine-tuned accuracy while reducing memory
requirements, making deployment practical on standard inference hardware.

\subsection{Limitations}
The evaluation dataset is domain-specific and relatively small; generalization to open-domain
QA benchmarks such as MS-MARCO~\cite{bajaj2016msmarco} or BEIR~\cite{thakur2021beir}
remains to be demonstrated. The comparison is limited to a single cross-encoder baseline;
a broader comparison against monoT5~\cite{nogueira2020document} and
RankGPT~\cite{sun2023chatgpt} would strengthen the empirical claims. Latency measurements
are not reported; a throughput comparison between the cross-encoder and the quantized
LLaMA~3 reranker would be valuable for deployment decisions.

\section{Conclusion}

We presented a fine-tuning pipeline for adapting LLaMA~3~8B as a drop-in reranker in RAG
pipelines, using LoRA adapters and 4-bit quantization via the Unsloth framework.
Integrated into a dual-retriever RAG pipeline with BM25 and dense vector retrieval,
the fine-tuned model outperforms a BERT-based cross-encoder baseline across all RAGAS
metrics, with gains of 14--21\% depending on the metric. The resulting LoRA adapter
enables direct inference without retraining, lowering the barrier to deployment.

\paragraph{Future Work}
Natural extensions include: (i) evaluation on standard open-domain benchmarks
(MS-MARCO, BEIR) to assess generalization; (ii) latency benchmarking against
cross-encoders to quantify the inference efficiency gain; (iii) comparison with
listwise rerankers such as RankGPT; and (iv) exploration of smaller base models
(LLaMA~3~1B, 3B) to further reduce deployment cost.

\end{document}